\documentclass[11pt,titlepage]{article}
\usepackage[margin=1.2in]{geometry}
\usepackage{amsmath, amsbsy, amsthm, array, mathrsfs}
\usepackage{graphics}
\usepackage{physics}
\usepackage{epsfig, amssymb,latexsym,verbatim}
\usepackage{graphicx}
\usepackage{color}
\usepackage{dsfont, bbm}
\usepackage{relsize}
\usepackage{subcaption}

\newcommand{\R}{\mathbb{R}}

\newcommand{\noin}{\noindent}
\newcommand{\bee}{\begin{eqnarray*}}
\newcommand{\ene}{\end{eqnarray*}}
\newcommand{\bec}{\begin{center}}
\newcommand{\enc}{\end{center}}
\newcommand{\be}{\begin{equation}}
\newcommand{\ee}{\end{equation}}

\newcommand{\mb}{\mathbf}
\newcommand{\bs}{\boldsymbol}
\newcommand{\tb}{\textbf}
\newcommand{\pend}{$\blacksquare$}
\newcommand{\vs}{\vskip 3mm}
\newcommand{\bi}{\begin{itemize}}
\newcommand{\ei}{\end{itemize}}

\begin{document}
\title{\LARGE  Non-asymptotic analysis and inference for an outlyingness induced winsorized mean
 \\[4ex]
}
\author{ {\sc Yijun Zuo}\\[2ex]
         {\small {\em  Department of Statistics and Probability,  Michigan State University} }\\
         {\small East Lansing, MI 48824, USA} \\
         {\small zuo@msu.edu}\\[6ex]
     }
 \date{\today}
\maketitle

\vskip 3mm
{\small

\begin{abstract}

Robust estimation of a mean vector, a topic regarded as obsolete in the traditional robust statistics community, has recently surged in machine learning literature in the last decade. The latest focus is on the sub-Gaussian performance and computability of the estimators in a non-asymptotic setting. Numerous traditional robust estimators are computationally intractable, which partly contributes to the renewal of the interest in the robust mean estimation.\vs

Robust centrality estimators, however, include the trimmed mean and  the sample median. The latter has the best robustness but suffers a low efficiency drawback.
Trimmed mean and median of means,
achieving sub-Gaussian performance have been proposed and studied in the literature.
\vs
This article investigates the robustness of leading  sub-Gaussian estimators of mean and reveals that none of them can resist greater than $25\%$ contamination in data and consequently  introduces an outlyingness induced winsorized mean which has the best possible robustness (can resist up to $50\%$ contamination without breakdown) meanwhile  achieving  high efficiency. Furthermore, it has a sub-Gaussian performance for uncontaminated samples and a bounded estimation error for contaminated samples at a given confidence level in a finite sample setting. It can be computed in linear time.

\bigskip
\par

\noindent{\bf Key words and phrase:} non-asymptotic analysis, centrality estimation, sub-Gaussian performance,  computability, finite sample breakdown point.
\bigskip
\par
\noindent {\bf Running title:} Outlyingness induced winsorized mean.
\end{abstract}
}
\setcounter{page}{3}

\section {Introduction}

Statisticians have recognized the
extreme sensitivity of the sample mean to unusual observations (outliers or contaminated points) and heavier-tailed data for a very long time.  In fact, robust alternatives of the sample mean were proposed and studied more than one century ago. Tukey's trimmed, winsorized mean, and weighted mean (and later Huber's M-estimator of location) are among popular univariate robust alternatives.  Arguably, the spatial median that dates back to Weber (1909) was the first robust alternative in high dimensions.\vs

In the last several decades, numerous multi-dimensional highly robust location estimators were proposed and examined in the traditional
 robust statistics community,
 including, among others,  the minimum volume ellipsoid (MVE) and the minimum
covariance determinant (MCD) estimator (Rousseeuw 1984,
1985), S estimates ( Rousseeuw and Yohai
(1984), Davies 1987), Stahel-Donoho estimators (“outlyingness” weighted mean) (Stahel (1981) and Donoho (1982), Zuo, et al (2004)), and many depth weighted and maximum depth estimators (see Zuo (2006a) for a survey). See Hubert, et al (2008) for a review on robust location estimators.
\vs
These alternatives, albeit being robust, lack computability in general (progress has been made in the literature though), which becomes the most vulnerable point of these estimators in real practice.  Constructing robust yet computable
mean estimators with sub-Gaussian performance (defined below) in high dimensions renews researchers great interest in ``big data"-era  in machine learning and other fields (see, Lerasle (2019), Diakonikolas and Kane (2019), Lecu\'{e} and Lerasle (2020), and Lugosi and Mendelson (2021), among others), where data are typically huge, have heavy-tails, and contain outliers (or contaminations).   
\vs
For a given 
 sample $\mb{X}^{(n)}:=\{\mb{X}_1, \cdots, \mb{X}_n\}$ from $\mb{X} \in \R^d$ ($d\geq 1$) with mean (or \emph{more generally location}) parameter $\bs{\mu}$ and covariance matrix $\mbox{Cov}(\mb{X})=\bs{\Sigma}$. Desired properties of any estimator  $\mb{T}(\mb{X}^{(n)})$ of $\bs{\mu}$  include \tb{(i)-(iv)} below, among others:

\vs
  \tb{(i)Affine equivariance}. Namely, $\mb{T}(\mb{A}\mb{X}^{(n)}+\mb{b})=\mb{A}\mb{T}(\mb{X}^{(n)})+\mb{b}$, for any non-singular $\mb{A}\in \R^{d\times d}$ and $\mb{b} \in \R^d$, where $\mb{A}\mb{X}^{(n)}+\mb{b}=\{\mb{A}\mb{X}_1 +\mb{b}, \cdots,\mb{A}\mb{X}_n +\mb{b} \}$. \vs

  In other words, the estimator does not depend on the underlying coordinate system and measurement scale. When the relation holds just for identity matrix $\mb{A}$, then the estimator is called \emph{translation} equivariant, whereas if $\mb{A}$ is an orthogonal matrix, then it is called \emph{rigid-body} equivariant. 
  The univariate sample median, defined below for a given univariate sample $z^{(n)}:=\{z_1,\cdots, z_n\}$, is affine equaivariant,
  \be
 \widehat{\mu}_n (z^{(n)})=\mbox{median}(z_1, \cdots, z_n)= \frac{z_{(\lfloor (n+1)/2\rfloor)}+z_{(\lfloor (n+2)/2\rfloor)}}{2},\label{median.eqn}
 \ee
  where $z_{(1)}\leq z_{(2)}\leq \cdots, \leq z_{(n)}$ are ordered values of  $z_is$ and $\lfloor ~\cdot\rfloor$ is the floor function.
If it is defined as any other point
  in between $z_{(\lfloor (n+1)/2\rfloor)}$ and $z_{(\lfloor (n+2)/2\rfloor)}$,
  then,  besides lack of uniqueness and well definedness, it also violates this desired property.  \vs

 \tb{(ii) Strong robustness}.  Namely, the estimator can perform well for heavy tailed data or data even containing outliers and contaminated points.
  \vs
  A prevailing quantitative measure
of robustness of an estimator in finite sample practice
 is the finite sample breakdown point (FSBP) introduced by Donoho and Huber (1983) (DH83).
\vs
\noin
\tb{Definition 1.1} [DH83]
The finite sample \emph{replacement breakdown point} (RBP) of a translation equivariant estimator $\mb{T}$
at $\mb{X}^{(n)}$ in $\R^d$ is defined as
\[
\mbox{RBP}(\mb{T}, \mb{X}^{(n)})=\min_{1\leq m\leq n}\{\frac{m}{n}:
\sup_{\mb{X}^{(n)}_{m}}\|\mb{T}(\mb{X}^{(n)}_{m})-\mb{T}(\mb{X}^{(n)})\|=\infty\},
\]
where $\mb{X}^{(n)}_{m}$ denotes a corrupted sample from $\mb{X}^{(n)}$ by
replacing $m$ points of $\mb{X}^{(n)}$ with arbitrary $m$ points, and $\|\cdot\|$ stands for Euclidean norm.
Namely, the RBP of an estimator is the
minimum replacement fraction of the contamination which could drive
the estimator beyond any bound. 
It is readily seen that the RBP of the sample mean is $1/n$, whereas that of the univariate sample median is $\lfloor(n+1)/2\rfloor/n$, the best possible RBP for any translation equivariant estimators (see Lopuha\"{a} and Rousseeuw (1991) (LR91)). The higher the RBP, the more robust the estimator is. In terms of FSBP, an estimator is \emph{strong robust} if its RBP is much larger than $1/n$.
\vs
\tb{(iii) Sub-Gaussian performance}. Namely that for a $\delta \in (0, 1)$ 
\be
\|\mb{T}(\mb{X}^{(n)})-\bs{\mu}\|> r_{\delta} \label{sub-gaussian.eqn}
\ee
with probability at most $\delta$, where $r_{\delta}=O\left(\sqrt{\frac{\mbox{Tr}(\bs{\Sigma})}{n}}+\sqrt{\frac{2\lambda_{\max}\log(1/\delta)}{n}}\right)$,  $\mbox{Tr}(\bs{A})$ stands for the trace of matrix $\bs{A}$, and $\lambda_{\max}$ is the largest eigenvalue of $\bs{\Sigma}$, see (1.3) of Lugosi and Mendelson (2021) (LM21). 
See pages 11-12 of Pauwels (2020) for sub-Gaussian definition. One hopes that the $r_{\delta}$ is as small as possible for a given $\delta$ (which is related to the confidence level), and the probability that (\ref{sub-gaussian.eqn}) holds true for an arbitrary ball radius $r$ is in order of $O(e^{-ar^2})$ for some constant $a$.
In the literature, when $d=1$,
$r_{\delta}$ is typically replaced by \be c \sqrt{\mbox{Var}({X}) \frac{\log(1/\delta)}{n}},\ee for some positive constant c; see (1.1) of LM21. Here $\mbox{Var}({X})$ is implicitly assumed to be known. So is the $\bs{\Sigma} $ above.
\vs

\tb{(iv) Computability}. Namely that the estimator can be computed in polynomial time.
\vs
\vs
The sample mean is a sub-Gaussian estimator of $\bs{\mu}$  when $\mb{X}$ follows a Gaussian distribution, see Lugosi and Mendelson (2019) (LM19). However, in other non-Gaussian cases, especially with heavy-tailed data or data containing outliers and contaminated points (this often happens in the machine learning field, genetic and genomic analysis, financial data analysis), its performance deteriorates dramatically (due to its low RBP $1/n$).\vs

In the last decade, numerous alternatives of sub-Gaussian estimators of $\bs{\mu}$ were proposed in the literature. Are they robust as supposed to be? Do they possess all desirable properties? 
It turns out that none of those investigated in next section  possesses the best RBP, $\lfloor(n+1)/2\rfloor\big/n$.
\vs
The reminder of the article is organized as follows. Section 2 investigates the popular sub-Gaussian estimators in the univariate case with emphasis on robustness.
Section 3 introduces an outlyingness induced wisorized mean that is affine equivariant, has the best possible RBP and can be computed in linear time. Section 4 presents the main results on the non-asymptotic analysis of the winsorized mean, establishing the sub-Gaussian performance of the estimator. Unlike much practice in the literature, our non-asymptotic analysis admits samples with contamination.  Concluding remarks end the article in Section 5.

\section{Univariate sub-Gaussian estimators}

Popular univariate sub-Gaussian mean estimators in the literature include, among others, (I) Catoni' estimator (Catoni (2012)),
(II) median-of-means estimator proposed and studied in many papers including, among others, Lerasle and Oliveira (2011) (LO11) and LM19, and (III) trimmed mean in LM19 and LM21. \vs

\noin
(I) \tb{Catoni's estimator}.  Catoni (2012) pioneered the  non-asymptotic deviation study for mean and variance estimation. Its mean estimator achieves the sub-Gaussian performance and possesses the computability (see remarks 2.1 below). However, it lacks strong robustness. In fact, one bad point can ruin the estimator (i.e. the RBP is $1/n$). The latter is defined as a new M-estimator, the solution of an estimation equation:
\be
\sum_{i=1}^n \psi(\alpha (Y_i-\widehat{\mu}))=0, \label{catoni.eqn}
\ee
where $Y_i$, $(i=1,\cdots, n)$ is a given sample from random variable $Y \in \R$ with E$Y=\mu$ and Var$(Y)=\sigma^2$, and $\alpha$ is a positive tuning parameter and $\psi(\cdot)$ is defined as
\[
\psi (x) = \left\{
\begin{array}{ll}
\log(1+x+x^2/2),& x\geq 0,\\[1ex]
-\log(1-x+x^2/2), & x\leq 0.
\end{array}
\right.
\]
\vs
In Catoni (2012), $\psi$ is allowed to be any one in-between the positive and negative pieces above but always is non-decreasing and the choice above is the preferred (see page 1152). 
\vs
\noin
\tb{Claim 2.1:} \tb{(a)} The RBP of $\widehat{\mu}$ given in (\ref{catoni.eqn}) is $1/n$. \tb{(b)} The $\widehat{\mu}$ is not affine equivariant.
\vs
\tb{Proof:} \tb{(a)} Contaminating one point from $\{Y_1,\cdots, Y_n\}$, say $Y_1$ is replaced by $C$, an arbitrary number. Let $C\to \infty$.
then $\widehat{\mu}\to \infty$ since otherwise, on the left-hand side (LHS) of equation (\ref{catoni.eqn}), there are $(n-1)$ terms bounded and one term becomes unbounded whereas the summation of all terms is zero, which is impossible.
\vs
\tb{(b)}  $\widehat{\mu}$ in (\ref{catoni.eqn}) is apparently translation equivariant but not scale equivariant that is, when replacing $Y_i$ with $sY_i$
in (\ref{catoni.eqn}), $s\widehat{\mu}$ is not necessarily the solution of the resulting equation.
 \hfill \pend
\vs
\noin
\tb{Remarks 2.1} \vs
\bi
\item[(i)]
The arguments in  the proof of (a) above hold for any estimation equation with an unbounded $\psi$ (such as (a) squared loss objective function for numerous estimators including LASSO in Hastie, et al (2015) and its various variants, e.g. the bridge estimators in Weng, et al (2018), in the literature and (b) Huber loss objective function in Sun, et al (2020) for the adaptive Huber regression estimator).
The low RBP of $\widehat{\mu}$ could be improved by replacing the unbounded $\psi$ with a bounded one such as the so-called \emph{narrowest choice} in Catoni (2012). The resulting one will possess an improved RBP (the equation (\ref{catoni.eqn}) might have more than one solution though).
\vs
\item[(ii)] A slight modification of (\ref{catoni.eqn}) could also lead to a affine equivariant  $\widehat{\mu}$. Simply using a pre-estimated scale such as the median of absolute deviations from the sample median (MAD) and divide $Y_i-\widehat{\mu}$ in (\ref{catoni.eqn}) by MAD, then the resulting $\widehat{\mu}$
    is affine equivarant. 
     However, the side effect of this modification above renders the derivation in Catoni (2012) for the sub-Gaussian performance of original estimator  invalid. 
\vs
\item[(iii)] It seems that there is no discussions on the computation of Catoni's estimator. Here is a feasible algorithm, write $w_i=\psi(\alpha(Y_i-\widehat{\mu}))/(\alpha(Y_i-\widehat{\mu}))$, then one have
an estimation equation $\sum_iw_i(Y_i-\widehat{\mu})=0$, then the solution can be obtained by iteratively procedure with an initial estimate $\widehat{\mu}_0$ (which could be the sample median or mean). 
     \hfill \pend
\ei
\vs
\noin
(II) \tb{Median-of-means (MOM) }  Median of means has been proposed in different forms and studied in the literature,  see, e.g. Nemirovsky and Yudin (1983),  Jerrum et al. (1986),  Alon et al. (2002), Hsu (2010), Lerasle (2019),
  LO11, and LM19.\vs

The procedure calls for the partitioning of a given data set $\mb{X}^{(n)}=\{\mb{X}_1, \cdots, \mb{X}_n\}$ into $k=n/m$ 
groups $B_1, \cdots, B_k$ with
the size $m$ of each group at least $2$, that is $|B_i|\geq 2$.  Next step is to compute the sample mean with respect to (w.r.t.) each group and obtains $k$ means, $\mb{Z}_1,\cdots, \mb{Z}_k$. The final step calculates the median of the $k$ means: $\mb{Z}_1,\cdots, \mb{Z}_k$.\vs
In the univariate case, the median is unambiguous, it is the regular sample median as defined in (\ref{median.eqn}). However, in both LO11 and LM19 it is defined to be any point
in-between $z_{(\lfloor (k+1)/2\rfloor)}$ and $z_{(\lfloor (k+2)/2\rfloor)}$, which leads to an estimator that might not be affine equivariant.
\vs
In the multivariate case, there are many choices of medians. For example, the coordinate-wise median (CM) and the spatial median (SM) are two that have the best RPB and high computability (both can be computed in $O(n)$ time). However, they are just translation equivariant not affine equivariant; the spatial median is rigid-body equivariant though.
\vs
\noin
\tb{Claim 2.2}: The RBP of MOM in $\R^d ~ (d\geq 1)$ is no greater than $ \lfloor{(n/m+1)/2\rfloor}/n $, $m\geq 2$.
\vs
\tb{Proof:} Assume that the median involved has the best possible RBP $\lfloor(n+1)/2 \rfloor/n $ (see LR91).
That is, in order to break down the median one has to contaminate $\lfloor(n+1)/2 \rfloor$ points w.r.t. a data set with sample size $n$ (this is clearly true in the univariate case, in multivariate case it is also true for CM and SM (but these two lose the affine equivariance)).
After the initial partition of an $n$ points data set, we obtain $k=\lfloor n/m\rfloor$ mean points, $\mb{Z}_1, \cdots, \mb{Z}_k$, where $m\geq2$. Therefore, the RBP of
MOM is no greater than $\lfloor{(n/m+1)/2\rfloor}/n $, $m\geq 2$. \hfill \pend
\vs
That is, the RBP of MOM in $\R^d$~(\tb{$\bs{d\geq 1}$})  can never be as high as $\lfloor(n+1)/2 \rfloor/n $.
\vs
\noin
(III) \tb{Trimmed mean in LM19, LM21}.  In both LM19 and LM21, a truncation function is defined as
\[
\phi_{\alpha, \beta}(x)=\left\{
\begin{array}{ll}
 \beta,& x>\beta,\\
 x,& x\in[\alpha, \beta],\\
 \alpha, & x<\alpha.
\end{array}
\right.
\]
Both are given two independent samples, $X_1,\cdots, X_n$ and $Y_1, \cdots, Y_n$ from $X$. $\alpha \leq \beta$ are estimated by the
ordered values $Y^*_1\leq Y^*_2,\cdots, \leq Y^*_n$ of $Y_1, \cdots, Y_n$. $\alpha=Y^*_{\varepsilon n}$ and $\beta=Y^*_{(1-\varepsilon)n}$ (assume that $\varepsilon n$ is an integer).
In LM19 and LM21, the trimmed mean is defined as
\be
\widehat{\mu}_n=\frac{1}{n}\sum_{i=1}^n\phi_{\alpha, \beta}(X_i).\label{trim.eqn}
\ee
\vs
In the traditional robust statistics, this type of estimator is called  the \emph{winsorized mean} since the data points on the two tails are modified, not trimmed/removed (as in the regular trimmed mean case).
Note that $\widehat{\mu}_n$ is not affine equivariant based on the current definition since the trimming is not symmetrically. But this can be easily fixed.
\vs
In LM21, samples are considered more realistically to be contaminated (with contamination rate of $\eta$) and $\alpha$, $\beta$, and $\widehat{\mu}_n$ are obtained based on the contaminated samples $\widetilde{X}_1,\cdots, \widetilde{X}_n$ and $\widetilde{Y}_1, \cdots, \widetilde{Y}_n$.
LM19, on the other hand, treated the idealized  non-contamination samples.

\vs

\noindent
\tb{Claim 2.3} The RBP of the $\widehat{\mu}_n$ in LM19, LM21 is  $\leq {(\lfloor \varepsilon n\rfloor}+1)/{2n}$, where $\varepsilon <1/2$.
\vs
\tb{Proof}: We show that the trimmed mean in (\ref{trim.eqn}) can be forced to be unbounded if we corrupt $\lfloor \varepsilon n\rfloor+1$ points among the give
uncontaminated $2n$ points: $X_1,\cdots, X_n$ and $Y_1, \cdots, Y_n$.
\vs
Replacing $\lfloor \varepsilon n\rfloor$ points in $\{Y_i\}$ with $M$, and one point in $\{X_i\}$ with $M-1$ and let $M\to -\infty$. It is readily seen that
$\alpha= M\to -\infty$ and so does $\widehat{\mu}_n$. \hfill \pend
\vs
That is, the RBP of the trimmed mean $\widehat{\mu}_n$ of LM19 and LM21 can never be as high as $\lfloor(n+1)/2 \rfloor/n $.
Now we summarize all results obtained so far with a table (Table \ref{RBP.table}).
\vs
\bec
\begin{table}[t!]
\centering {Performance summary of three leading estimators}\\[1ex]
\begin{tabular} {c c c c c}
estimator & affine equivariant & RBP  & sub-Gaussian & computability\\[2ex]
\hline\\[1ex]
Catoni's & Not (fixable) & $\frac{1}{n}$ (improvable)& Yes (conditionally) & Yes\\[2ex]
MOM & Not (fixable) & $ \leq \frac{(n/m+1)/2}{n}, m\geq 2 $& Yes (conditionally) & Yes\\[2ex]
Trimmed mean & Not (fixable) & $\leq \frac{\lfloor(\varepsilon n\rfloor+1}{2n}, \varepsilon<\frac{1}{2} $& Yes (conditionally)& Yes\\[2ex]
\hline\\[1ex]
\end{tabular}
\vspace*{-7mm}
\caption{Performance of three leading sub-Gaussian estimators of mean w.r.t. four criteria} \label{RBP.table}
\end{table}
\enc
\vspace*{-5mm}
\noindent
\tb{Remarks 2.2}
\bi
\item[(i)]
 Based on the table \ref{RBP.table}, if we let $n\to \infty$ in the three RBPs, then we obtain asymptotic breakdown points (ABP). None of the ABP is greater than $25\%$. That is, none of the three estimators can resist greater than $25\%$ contamination  asymptotically.
 \item[(ii)] The sub-Gaussian results for the three estimators are conditional since it is implicitly assumed that no contamination in a sample.
If considering the possible contamination in a real sample, then the sub-Gaussian performance analysis in Catoni (2012) for Catoni's estimator, in Lerasle (2019), LO11 and LM19 for MOM estimator, in LM19 for the trimmed mean, is invalid. The non-asymptotic result obtained in LM21 for the trimmed mean is more realistic than that of LM19 since it considers the possible contamination in data. Unfortunately, the contamination rate $\eta$ in LM21 is implicitly required to be less than $6.25\%$, 
 which is not realistic for a real dataset.
\ei
\vs
Next we introduce an estimator for centrality that has the best RBP: $\lfloor(n+1)/2 \rfloor/n $ (i.e. ABP is $50\%$). Our non-asymptotic analysis can handle contamination rate up to $25\%$ in a sample, in sharp contrast to $6.25\%$ in LM21.
\vs

\vs
\section{An outlyingness induced winsorized mean}
Let $\mu(F)$ and $\sigma(F)$ be some robust location and scale measures of
a distribution $F$ of random variable $X$. For simplicity and robustness, we consider $\mu$ and $\sigma$ being the
median (Med) and the median of absolute deviations (MAD) throughout. Assume $\sigma(F) > 0$, namely, $F$ is not degenerate.
For a given point $x \in \R^1$, we define the \emph{outlyingness of x} w.r.t. the centrality of $F$ (see Stahel (1981) and Donoho (1982) for a high dimensional version) by
\be
 O(x, F)=|x-\mu(F)|/\sigma(F). \label{outlyingness.eqn}
\ee
It is readily seen that $O(x, F)$ is a generalized standard deviation, or equivalent to the one-dimensional projection depth (see Zuo and Serfling(2000) and Zuo (2003) for a high dimensional version). Let $\beta>0$ be a fixed number (or outlyingness), define
$$
L(F)=\mu(F)-\beta\sigma(F);~~ U(F)=\mu(F)+\beta\sigma(F).
$$
We have suppressed $\beta$ above, sometimes we will suppress $F$ as well. Define  functions
\begin{align}
\phi(x):=\phi_{L, ~U}(x, F)&=L\mathds{1}(x<L)+x\mathds{1}(L\leq x\leq U)+U\mathds{1}(x>U), \label{phi.eqn}  \\[2ex]
T^{\beta}_w(F)&= \frac{\int \phi(x)w(O(x,F))dF(x)}{\int w(O(x,F))dF(x)}, \label{winsorized.eqn}
\end{align}
where $0\leq \beta\leq \infty$ and $w$ is a bounded weight function on $\R^1$ so that the denominator is positive. Note that $T^{\beta}_w(F)$ here is exactly the winsorized mean defined in (2.3) of Wu and Zuo (2009) (WZ09) and is closely related to the one-dimensional special case of projection depth trimmed mean studied in Zuo (2006b).
\vs
 When $w(x)=c > 0$,  in the extreme case that $\beta=0$, then $T^{\beta}_w(F)$ is the median functional, in the case that $\beta=\infty$, then it becomes the mean functional. In the sequel, we consider $w(x)=c > 0$ and a fixed $0<\beta<\infty$ (a special case that has never been introduced and studied in non-asymptotic sense before), that is
 \be
 T(F):= T^{\beta}_w(F)=\int\phi_{L,~ U}(x) dF(x)=E(\phi_{L,~U}(X)). \label{winsorized1.eqn}
 \ee
 {\it Note that the mean functional does not always exist while $T(F)$ does}.  {\it If E$(X)$ exists and $F$ is symmetric, then $T(F)$ and $E(X)$ are identical.}  Consequently, it makes more sense to estimate the $T(F)$ as a measure of centrality parameter.\vs
 When replacing $F$ with its empirical version $F_n$, it is readily seen that
 \be
 T(F_n)=\int \phi_{L_n, ~U_n}(x) dF_n(x), \label{sample.eqn}
 \ee
 is a winsorized mean which replaces sample points below $L_n$ by $L_n$ and above $U_n$ by $U_n$ and keeps the points in-between untouched. It then takes an average of resulting $n$ points, where $L_n:=L(F_n)$ and $U_n:=U(F_n)$.
\vs
The sample mean treats each sample point equally important and therefore suffers a low breakdown point since a single outlier on a tail can ruin it.
Traditional trimmed mean and sample median are robust alternatives as the estimator of centrality parameter.  The former truncates the same number of data points at the two tails of the data cloud, the latter even truncates all the left and right points to the median (the average of middle two points). Both lose some information about the data.
\vs
$T(F_n)$ is different from  the trimmed mean, the median and even the regular winsorized mean in that it does not necessarily replace the same number of points at both tails (see WZ09 for more discussion).
$T(F_n)$ certainly also can serve as estimator for centrality. It possesses the best possible breakdown point with a RBP $\lfloor (n+1)/2\rfloor\big/n $ and can be very efficient (see WZ09). It is affine equivariant (see page 135 of WZ09). Large sample properties of its general version have been investigated in WZ09, but its finite sample behavior (other than FSBP) has never been explored. In the following, we study the accuracy and confidence of the estimation of $T(F_n)$ for $T(F)$ in a non-asymptotic setting.

\section{Main results}
Assume that  $X_1,\cdots, X_{n}$ from $F$ of $X$ is an uncontaminated i.i.d sample. 
In  reality  we might just have a contaminated sample $X^c_1, \cdots, X^c_{n}$ which contains $\varepsilon n$ (assume it is an integer for simplicity) arbitrarily contaminated points with $0\leq\varepsilon n<\lfloor(n+1)/2\rfloor$ (somewhat like in a {Huber $\varepsilon$ contaminated model} or in a RBP contamination setting). Denote by $L^c_n$ and $U^c_n$ to signify that the two depend on the contaminated sample.
\vs
We want to assess the estimation error (accuracy), or equivalently, to provide a sharp upper bound for the deviation below with a great confidence: $1-\delta$, for a given $\delta \in (0, 1)$
\be
\Big|\int \phi_{L^c_n, ~U^c_n}(x) dF^c_n(x)-\int \phi_{L,~U}(x) dF(x)\Big|
=\Big|\frac{1}{n}\sum_{i=1}^n \phi_{L^c_n,~U^c_n}(X^c_i)-T(F)\Big|.
 \ee
 \vs
Let $F_Y^{-1}(p):=\inf\{x:, F_Y(x)\geq p\}$ for any random variable $Y\in \R^1$. 
We assume that: \vs 
 \tb{(A1)}:
 The variance of $X$ exists, that is, $\sigma^2_X<\infty$.
\vs
\tb{(A2)}: For small $\alpha\in(0, c)$, (i) $|F^{-1}_X(1/2\pm \alpha)-F^{-1}_X(1/2)|\leq C_1\alpha$, (ii) $|F_Z^{-1}(1/2\pm \alpha)-F^{-1}_Z(1/2)|\leq C_2\alpha$, where $Z=|X-F_X^{-1}(1/2)|$, $c<1/2$, $C_1$ and $C_2$ are positive constants.
\vs
Clearly, sufficient conditions for \tb{(A2)} to hold true include an $F_X$ with a positive continuous density around its median and MAD points.
\vs
Before presenting main results, we cite an important concentration inequality 
 which we will employ repeatedly in the following proofs.
\vs
\vs
\noindent
\tb{Lemma 4.0 (Bernstein's inequality)} [Boucheron, et al (2013); 
Bernstein (1946)]
Let $X_1,\cdots, X_n$ be independent random
variables with finite variance such that $X_i \leq b$ for some $b > 0$ almost surely
for all $i\leq n$ and $E(X_i)=0$. Let $\nu=\frac{1}{n}\sum_{i=1}^n\mbox{Var}(X_i)$.  Then for any $t>0$, we have
\[
P\left(\frac{1}{n}\sum_{i=1}^nX_i\geq t \right)\leq \exp\left(-\frac{nt^2}{2(\nu+bt/3)}\right).
\]
\vs
\noin
Identical distribution of $X_i$ is not required above, only independence is necessary. \hfill \pend
\vs
\noindent
\tb{Lemma 4.1} Under \tb{(A2)}, for a given $\delta \in (0,1)$  with probability at least $1-\delta/3$, we have that (i) $|\mu(F_n^c)-\mu(F)|\leq C_1\varepsilon^*$ and (ii) $|\sigma(F_n^c)-\sigma(F)|\leq C_2\varepsilon^*/\beta$ provided that $\delta>12 \delta^*$, where $\varepsilon^* \in [2\varepsilon, 1/2)$, 
and $\delta^*:=\exp{-n(\varepsilon^*)^2/(8(1/4-(\varepsilon^*)^2+\varepsilon^*/6))}$.
\vs
The Lemma claims that the median and MAD obtained based on the contaminated sample (with contamination rate $\varepsilon$) are still very close to their true population version, respectively.
\vs
\noindent
\tb{Proof}: It suffices to show that with probability at least $1-\delta/6$ that $|\mu(F_n^c)-\mu(F)|\leq C_1\varepsilon^*$. (ii) could be treated similarly.\vs
A straightforward application of Bernstein's inequality 
 yields that with probability at least 
$1-\delta^*$
\be
\sum_{i=1}^n \mathds{1}\{X_i\leq F^{-1}_X(1/2+\varepsilon^*)\}>(1/2+\varepsilon^*/2)n,
\ee
 and with probability at least $1-\delta^*$
\be
\sum_{i=1}^n \mathds{1}\{X_i> F^{-1}_X(1/2-\varepsilon^*)\}>(1/2+\varepsilon^*/2)n.
\ee
Since there are at most $\varepsilon n $ points which are contaminated among the $n$ point of $\{X_i\}$, so we have that
with probability at least $1-2\delta^*$
\be
\sum_{i=1}^n \mathds{1}\{X^c_i\leq F^{-1}_X(1/2+\varepsilon^*)\}>(1/2+\varepsilon^*/2)n-\varepsilon n \geq n/2,
\ee
 and 
\be
\sum_{i=1}^n \mathds{1}\{X^c_i> F^{-1}_X(1/2-\varepsilon^*)\}>(1/2+\varepsilon^*/2)n-\varepsilon n\geq n/2.
\ee
The last two expressions imply that with probability at least $1-2\delta^*$
\be
 F^{-1}_X(1/2-\varepsilon^*) \leq \mu(F_n^c) \leq F^{-1}_X(1/2+\varepsilon^*).
\ee
This, in conjunction with \tb{(A2)}, implies that $|\mu(F_n^c)-\mu(F)|\leq C_1\varepsilon^*$ with probability
at least $1-\delta/6$ as long as $\delta> 12\delta^*  =12 \exp{-n(\varepsilon^*)^2/(8(1/4-(\varepsilon^*)^2+\varepsilon^*/6))}$.
\vs
Now for (ii), we treat $\varepsilon^*/\beta$ as $\varepsilon^*$ in (i), and everything is similar. Note that $\delta^*$ is a non-increasing function of $\varepsilon^*$ so when $\beta\geq 1$, (ii) holds with probability $1-\delta/6$ as long as $\delta>12 \exp{-n(\varepsilon^*)^2/(8(1/4-(\varepsilon^*)^2+\varepsilon^*/6))}$. \hfill \pend
\vs
\noindent
\tb{Remarks 4.1}\bi
\item[{(i)}] It is readily seen that if we set $\varepsilon^*$ to be its smallest value $2 \varepsilon$, then the proof above and Lemma 4.1 hold. For generality, we keep using $\varepsilon^*$ in the sequel.
\item[{(ii)}] The Lemma holds true for any contamination rate $\varepsilon$ up to $25\%$. 
\hfill \pend
\ei
\vs
\noindent
\tb{Corollary 4.1} Under \tb{(A2)}, with the $\varepsilon^*$ and $\delta$ in Lemma 4.1, we have that $|L_n^c-L|\leq C\varepsilon^*$ and $|U_n^c-U|\leq C\varepsilon^*$  with probability at least
$1-\delta/3$, where $C=C_1+ C_2$.
\vs
\noindent
 \tb{Proof}: This follows straightforwardly from Lemma 4.1, details are skipped. \hfill \pend

\vs
\noindent
\tb{Lemma 4.2}.  Under \tb{(A2)}, with the $\varepsilon^*$ and $\delta$ in Lemma 4.1, we have that
\be
\Big| \frac{1}{n}\sum_{i=1}^n\Big(\phi_{L^c_n,~U^c_n}(X^c_i)-\phi_{L^c_n,~U^c_n} (X_i)\Big)\Big|\leq \varepsilon(2\beta\sigma+2C\varepsilon^*),\label{diff.eqn}
\ee
with probability at least $1-\delta/3$, where $C$ is given in Corollary 4.1.
\vs
\noindent
\tb{Proof}: On the LHS of (\ref{diff.eqn}), there are  at most $\varepsilon n$ terms that are non-zero. For each non-zero term, the maximum difference between $\phi_{L^c_n,~U^c_n}(X^c_i)$ and $\phi_{L^c_n,~U^c_n} (X_i)$ is no greater than $|U^c_n-L^c_n|\leq |U+C\varepsilon^*-(L-C\varepsilon^*)|=2\beta\sigma+2C\varepsilon^*$ with probability at least $1-\delta/3$ in light of Corollary 4.1. This completes the proof. \hfill \pend

\vs
\noindent
\tb{Lemma 4.3} Under \tb{(A2)}, with the $\varepsilon^*$ and $\delta$ in Lemma 4.1, we have that
\be
\Big| \frac{1}{n}\sum_{i=1}^n\Big(\phi_{L^c_n,~U^c_n}(X_i)-\phi_{L,~U} (X_i)\Big)\Big|\leq C\varepsilon^*,
\ee
with probability at least $1-\delta/3$, where $C$ is given in Corollary 4.1.
\vs
\noindent
\tb{Proof}: This follows straightforwardly from Corollary 4.1. \hfill\pend
\vs
\noindent
\tb{Lemma 4.4}  Under \tb{(A1)-(A2)}, 
 we have that
\be
 \Big| \frac{1}{n}\sum_{i=1}^n\Big(\phi_{L,~U} (X_i)- E\big(\phi_{L,~U} (X_i)\big)\Big) \Big|
 \leq \sqrt{2}\left( \frac{U^*}{\sigma_X} \sqrt{ \frac{2\log 6/\delta}{n} }+1\right) \sigma_X \sqrt{ \frac{\log 6/\delta}{n}}, \label{diff2.eqn}
\ee
with probability at least $1-\delta/3$, where $U^*:=|\mu|+\beta\sigma$.
\vs
\noindent
\tb{Proof}: Note that the LHS of (\ref{diff2.eqn}) is the sum of i.i.d. centered random variables which are bounded by $2U^*$ with variance at most $\sigma_X^2$. Consequently, by Bernstein's inequality we have
\bee
\frac{1}{n}\sum_{i=1}^n\Big(\phi_{L,~U} (X_i)- E\big(\phi_{L,~U} (X_i)\big)\Big) &\leq &\frac{2U^*}{n}\log 6/\delta +\sigma_X \sqrt{\frac{2\log6/\delta}{n}},
\ene
with probability at least $1-\delta/6$.
\vs
An identical treatment for the other tail yields the desired result. \hfill \pend
\vs
Now we are in the position to present the main results.

\vs
\noindent
\tb{Theorem 4.1} Under \tb{(A1)-(A2)} with $C$ and $\varepsilon^*$ given in Lemma 4.1, we have
\be
\Big|T(F^c_n)-T(F)\Big|\leq \sqrt{2}\left( \frac{U^*}{\sigma_X} \sqrt{ \frac{2\log 6/\delta}{n} }+1\right) \sigma_X \sqrt{ \frac{\log 6/\delta}{n}}
+C\varepsilon^*+\varepsilon(2\beta\sigma+2C\varepsilon^*),
\ee
with probability at least $1-\delta$, provided that $\delta>12\exp{-n(\varepsilon^*)^2/(8(1/4-(\varepsilon^*)^2+\varepsilon^*/6))}$.
\vs
\noindent
\tb{Proof}: First we write
\bee
\Big|T(F^c_n)-T(F)\Big|&=&\Big|\frac{1}{n}\sum_{i=1}^n \phi_{L^c_n,~U^c_n}(X^c_i)-T(F)\Big|\\
&\leq&\Big|\frac{1}{n}\sum_{i=1}^n\Big(\phi_{L^c_n,~U^c_n}(X^c_i)-\phi_{L^c_n,~U^c_n} (X_i)\Big)\Big|\\
&&+\Big| \frac{1}{n}\sum_{i=1}^n\Big(\phi_{L^c_n,~U^c_n}(X_i)-\phi_{L,~U} (X_i)\Big)\Big|\\
&&+ \Big| \frac{1}{n}\sum_{i=1}^n\Big(\phi_{L,~U} (X_i)- E\big(\phi_{L,~U} (X_i)\big)\Big) \Big|,
\ene
the desired result follows straightforwardly from Lemmas 4.4, 4.3 and 4.2.
\hfill \pend
\vs
\noindent
\tb{Remarks 4.2}
\bi
\item[(i)] The parameters in the upper bound in Theorem 4.1 include $\sigma_X$, $\mu$, $\sigma$, $C$. These all depends on 
$F_X$. $\delta$ is a selected number and related to the desired confidence level.  $\varepsilon$ is the contamination level which is allowed to be up to $25\%$ (unlike in LM21, it is restricted to be less than $6.25\%$).
    $\varepsilon^*$ can be set to be $2\varepsilon$. $\beta\geq 1$ is a pre-selected constant.

\item[(ii)] If we assume that there is no contamination in the data like many articles on sub-gaussian estimators in the literature, then Theorem 4.1 gives clearly the sub-Gaussian error rate.
In general, if $C\varepsilon^*+\varepsilon(2\beta\sigma+2C\varepsilon^*)$ is $O\Big( \sigma_X \sqrt{ \frac{\log 6/\delta}{n}})\Big)$, then Theorem 4.1 again gives the sub-Gaussian error rate.

 \item[(iii)] One might be more conformable if the $T(F)$ in the theorem is replaced by $E(X)$. $T(F)$ is closely related to the $E(X)$. In fact, if  $\beta \to \infty$, then $T(F)\to E(X)$. But when $\beta\to \infty$, Lemma 4.1 might not hold (alternatively, one can choose large enough $\beta$, see Theorem 4.2 below). On the other hand,  the mean of any distribution $F_X$ does not always exist while $T(F_X)$ does. If $m=E(X)$ exists and $F$ is symmetric, then $T(F)=m$. For a general non-symmetric $F$, we have the following.
  \hfill \pend
\ei
\vs
\noin
\tb{Theorem 4.2} Assume that $m=E(X)$ exits and $F$ has a positive density. Under \tb{(A1)-(A2)} with $C$ and $\varepsilon^*$ given in Lemma 4.1, let $\delta \in (0,1)$ such that $\delta \geq 6 \exp{-n}$ and let $\eta={ \frac{\log 6/\delta}{n}}$, and
$\sigma \beta=\max\{{F^{-1}_{X-m}(1-\eta/2)+m-\mu}, {\mu-m-F^{-1}_{X-m}(\eta/2)} \}$. 
 we have
\be
|T(F^c_n)-E(X)|\leq \sqrt{2}\left( \frac{U^*}{\sigma_X} \sqrt{ \frac{2\log 6/\delta}{n} }+3 
\right) \sigma_X \sqrt{ \frac{\log 6/\delta}{n}}
+C\varepsilon^*+\varepsilon(2\beta\sigma+2C\varepsilon^*),
\ee
with probability at least $1-\delta$, provided that $\delta>12\exp{-n(\varepsilon^*)^2/(8(1/4-(\varepsilon^*)^2+\varepsilon^*/6))}$.
\vs
\noin
\tb{Proof}: By (\ref{phi.eqn}) and (\ref{winsorized1.eqn}), it is readily seen that 
\begin{align}
|T(F)-E(X)|&=\big|E\big[(X-L)\mathds{1}(X<L)+(X-U)\mathds{1}(X>U)  \big]\big|\nonumber \\
&\leq \big|E(X-L)\mathds{1}(X<L) \big|+ \big|E(X-U)\mathds{1}(X>U) \big|.\label{triangle.eqn}
\end{align}
Let us first treat $ \big|E(X-U)\mathds{1}(X>U) \big|$, the first term on the RHS can be treated similarly. Write $M:=U-m=F^{-1}_{X-m}(1-\eta/2)$, where $\eta={ \frac{\log 6/\delta}{n}}\leq 1$.
Hence, 
\be\frac{\eta}{2}=P(X>U)\leq\frac{\sigma^2_X}{M^2}. \label{quantile.eqn}
\ee
Now, by H\"{o}lder inequality we have
\begin{align}
\big|E(X-U)\mathds{1}(X>U) \big| &\leq \big|E(X-m)\mathds{1}(X>U)\big|+\big|EM\mathds{1}(X>U)\big|\nonumber\\
&\leq \sigma_X\sqrt{P(X>U} +|M|P(X>U)\nonumber\\
&\leq \sigma_x\sqrt{\eta/2}+ \sigma_x\sqrt{\eta/2}\nonumber\\
&=\sigma_x\sqrt{2\eta},\nonumber
\end{align}
where the last inequality follows from (\ref{quantile.eqn}). In the above we have assumed that $M\neq 0$, if it is zero, then upper bound above still holds. 
Set $$\beta_U=\frac{F^{-1}_{X-m}(1-\eta/2)+m-\mu}{\sigma}.$$ Likewise to treat $\big|E(X-L)\mathds{1}(X<L) \big|$ we can set $$\beta_L=\frac{\mu-m-F^{-1}_{X-m}(\eta/2)}{\sigma},$$ and obtain
 $$\big|E(X-L)\mathds{1}(X<L) \big|\leq \sigma_x\sqrt{2\eta}.$$
Finally, let $\beta=\max\{\beta_U, \beta_L\}$. These, in conjunction with Theorem 4.1 and (\ref{triangle.eqn}), complete the proof.
\hfill \pend
\vs
\noindent
\tb{Remarks 4.3}
\bi
\item[(i)] All remarks on the sub-gaussian error rate in Remarks 4.2 are still valid in this case with the $T(F)$ being replaced by $E(X)$.

\item[(ii)] Since $m=E(X)$ is unknown, so we can replace $\mu-m$ in the definition of $\beta$ by $\sigma_X$ since $|\mu-m|\leq \sigma_X$. That is, $\sigma \beta=\max\{{F^{-1}_{X-m}(1-\eta/2)+\sigma_X}, ~ {\sigma_X-F^{-1}_{X-m}(\eta/2)} \}.$ \hfill \pend
\ei
\section{Concluding Remarks}
\tb{Non-asymptotic versus asymptotic analysis}.  With the assistance of the beautiful (and powerful) probability theory,  asymptotic analysis or large sample study of estimators has dominated modern statistics research since its beginning. Recently, computer scientists and statisticians in machine learning and data mining have renewed interest in centrality estimation to assist their analysis of ``big data". Because of large but finite sample sizes, these people have shifted their interest to non-asymptotic analysis.\vs
 In fact, this type of analysis, complementary to the asymptotic one,  is important for any estimator, not only for the mean estimator. Here researchers are interested in the tail probabilities of the distribution of the estimation error (aka deviation (from the target)).
  They hope that probability mass concentrating around the target is as high as possible and the radius of the ball centered at the target  is as small as possible for a fixed level of probability (confidence). Ideally, the tail probabilities (or the probabilities out side the ball)
 decrease exponentially when the radius of the ball increases. Such a distribution of the estimation error 
 is called sub-Gaussian (see pages 11-12 of Pauwels (2020), or page 13 of Lerasle (2019)); the estimator involved has a sub-Gaussian performance. The latter has been enthusiastically pursuing in the machine learning/statistics
 literature for the mean vector.
\vs
\noin
\tb{Centrality or more general location measure, Why always the sample mean?} 
Academics prefer the sample mean partially because it has some optimal properties (especially in the normal model case, see Proposition 6.1 of Catoni (2012)), e.g., it is the minimizer of the empirical squared error loss, and partially due to the law of large numbers and the central limit theorem.
\vs
Realizing the extreme sensitivity of the sample mean to a slight normality departure, such as a heavy-tailed (or contaminated and even containing outliers) data set, statisticians introduced trimmed, weighted/winsorized mean, and median as robust centrality or location estimating alternatives. 
 Unlike the expectation, they always exist.
 Their sample versions estimate the population counterparts respectively. The latter might not be the expectation (they are identical only when the underlying distribution is symmetric) but certainly can serve as a measure for the centrality (or location) parameter. This motivates us to address the winsorized mean which is more robust than the one given in LM19 or LM21.\vs
  \noin
\tb{Assumptions on $F_X$}.
In the literature for the sub-Gaussian estimators of the mean, $\sigma_X$ is usually assumed to be known (for the sake of applying Chebyshev's inequality). However, the first moment is unknown and we are estimating it, assumptions surrounding $\sigma_X$ are likely to be unsound in practice. Of course, the assumption \tb{(A1)-(A2)} here are also not realistic.\vs
\noin
\tb{Relevance of affine equivariance and RBP}. Some people might question
 the relevance of these two performance criteria. Affine equivariance guarantees that the median temperature day  of Fairbanks, Alaska in January is the same day no matter one uses the Fahrenheit ($^\circ $F) scale, or the Celsius ($^\circ$C) scale. Non-asymptotic results can not replace that of the RBP, one measures the performance of an estimator (w.r.t. estimation error or accuracy) and the other measures the resistance to the effect (on the estimator) of the contamination, outliers, or points from the heavy tailed distribution. Both are in a finite sample setting but irreplaceable with each other.
\vs
\noin
\tb{Computability.} 
There is no problem with the computation of one-dimensional sub-Gaussian estimators, all in this article can be computed in linear time.
Intractability of computation appears when one extends the univariate median 
to a multivariate version in the MOM in high dimensions.
Fortunately, promising non-asymptotic results on the two leading depth induced multi-dimensional medians, the halfspace and the projection medians, have been obtained recently by Chen, Gao and Ren (2018) and Depersin and Lecu\'{e} (2021), respectively. These two depth medians possess nice properties such as affine equivariance and high breakdown point (Zuo (2003)). Their computation has to be carried out approximately and has been addressed in Rousseeuw and Ruts (1998) for the bivariate halfspace median  in Liu (2017) for the projection median  in dimension $d>3$. High dimensional halfspace median could be computed approximately using the new approach for depth calculation in Zuo (2018).
\vs
\noin
\tb{Some open and future problems}. Although (\ref{winsorized1.eqn}) is a special case of (\ref{winsorized.eqn}), but it has never been introduced and studied before. 
To extend our study here to its high dimensional version seems intriguing yet challenging.
Non-asymptotic performance analysis of any estimators in high dimensions without the assumption that $\mb{\Sigma}$ is known and allowing the contamination of data is still an open problem (progress has been made in the literature though).

\vs
\begin{center}
{\textbf{\large Acknowledgments}}
\end{center}
\vspace*{-2mm}
The author thanks Hanshi Zuo and Prof.s Wei Shao, Yimin Xiao, and Haolei Weng for 
insightful comments and stimulus discussions.
\vs

{\small

\begin{thebibliography}{9}
\bibitem{AMS02} Alon, N., Matias, Y., and Szegedy, M. (2002), ``The space complexity of approximating the frequency moments",
\emph{Journal of Computer and System Sciences}, 58:137–147.

\bibitem{B46} Bernstein, S.N., ”The Theory of Probabilities”, Gastehizdat Publishing
House, Moscow, 1946.
\bibitem{BLM13} Boucheron, S.,  Lugosi, G., and Massart, P. (2013), ``Concentration
Inequalities: A Nonasymptotic Theory of Independence'', Oxford University
Press, 2013. ISBN 978-0-19-953525-5.

\bibitem{O12}  Catoni, O. (2012), ``Challenging the empirical mean and empirical variance: a deviation study". \emph{Annales de
l’Institut Henri Poincaré, Probabilités et Statistiques}, 48(4):1148–1185.

\bibitem{CG18} Catoni, O., and I. Giulini, I. (2018), ``Dimension-free PAC-Bayesian bounds for the estimation of the mean of a
random vector", arXiv preprint arXiv:1802.04308.

\bibitem{CGR18} Chen, M., Gao, C. and Ren, Z. (2018), ``Robust covariance and scatter matrix estimation under Huber’s
contamination model'', \emph{Ann. Statist.} 46 1932–1960.

\bibitem{D87} Davies, P. L.  (1987), ``Asymptotic behavior of S-estimators of multivariate
location parameters and dispersion matrices", \emph{Ann. Statist.} 15 1269-1292

\bibitem{DL21} Depersin, J. and · Lecué, G. (2021), ``On the robustness to adversarial corruption and to heavy-tailed data of the Stahel-Donoho median of means", arXiv:2101.09117v1.

\bibitem{IL19} Diakonikolas,I., and Kane, D., (2019),  ``Recent Advances in Algorithmic High-Dimensional Robust Statistics", arXiv:1911.05911v1.
\bibitem{D82}  Donoho, D.\ L.\  ``Breakdown properties of multivariate location estimators". PhD Qualifying
 paper, Harvard Univ. (1982).

\bibitem{D/H} Donoho,\ D.\ L., and Huber,\ P.\ J. (1983), ``The notion of
breakdown point", in: P.\ J.\ Bickel, K.\ A.\ Doksum and J.\ L.\ Hodges, Jr., eds. {\it A Festschrift
foe Erich L.\ Lehmann} (Wadsworth, Belmont, CA) pp.\ 157-184.
\bibitem{HTW15} Hastie, T., Tibshirani, R., and Wainwright, M. J. (2015), ``Statistical Learning
With Sparsity: The Lasso and Generalizations'', Boca Raton, FL: CRC Press
\bibitem{H10} Hsu, D. (2010), ``Robust statistics", http://www.inherentuncertainty.org/2010/12/robust-statistics.html.
\bibitem{HRS08}Hubert, M., Rousseeuw, P. J., and Van Aelst, S. (2008), ``High-Breakdown Robust Multivariate Methods",
\emph{Statistical Science}, Vol. 23, No. 1, 92–119.

\bibitem{JVV86}Jerrum, M., Valiant, L., and Vazirani, V. (1986), ``Random generation of combinatorial structures from a uniform
distribution", \emph{Theoretical Computer Science}, 43:186–188.

\bibitem{L19} Lerasle, M. (2019)), ``Selected topics on robust statistical learning theory, Lecture Notes",arXiv:1908.10761v1.
\bibitem{LO11}  Lerasle, M. and Oliveira, R. I. (2011), ``Robust empirical mean estimators''. Preprint. Available at
arXiv:1112.3914.

\bibitem{l17} Liu, X., (2017), ``Approximating projection depth median of dimensions $p \geq 3$". \emph{Commun Stat-Simul C} 46, 3756-3768.
\bibitem{LR} Lopuha\"{a},\ H.\ P. and Rousseeuw,\ J. (1991), ``Breakdown
points of affine equivariant estimators of multivariate location and
covariance matrices", {\it Ann. Statist.} {\bf 19}, 229-248.

\bibitem{LL20} Lecu\'{e}, G. and Lerasle, M. (2020). ``Robust machine learning by median-of-means: Theory and practice."
{\it Ann. Statist.} 48 906–931.


\bibitem{LM19} Lugosi, G, and  Mendelson, S. (2019), ``Mean Estimation and Regression Under Heavy-Tailed
Distributions: A Survey", \emph{Foundations of Computational Mathematics}, 19:1145–1190.


\bibitem{LM20} Lugosi, G, and  Mendelson, S. (2021),
``Robust multivariate mean estimation: the optimality of trimmed mean".
\emph{Ann. Statist.},  Vol. 49, No. 1, 393–410,
https://doi.org/10.1214/20-AOS1961.


\bibitem{NY83}  Nemirovsky, A.S., and Yudin, D.B. (1983), Problem complexity and method efficiency in optimization.
\bibitem{P20} Pauwels, E., (2020), ``Lecture notes: Statistics, optimization and algorithms in high dimension", https://www.math.univ-toulouse.fr/\~{}epauwels/M2RI/.
\bibitem{R84} Rousseeuw,\ P.\ J. (1984), ``Least median of squares
regression", {\it J. Amer. Statist. Assoc.} {\bf 79}, 871-880.

\bibitem{R85} Rousseeuw, P. J. (1985). Multivariate estimation with high breakdown point.
In Mathematical Statistics and Applications (Eds. W. Grossmann, G. Pflug,
I. Vincze and W. Wertz). Reidel. 283-297.

\bibitem{RR98} Rousseeuw, P.J., and Ruts, I. (1998), ``Construting the bivariate Tukey median",
\emph{Statistica Sinica}, Vol. 8, No. 3, pp. 827-839.
\bibitem{RY84}
Rousseeuw, \ P.\ J.  and  Yohai, V. J. (1984). Robust regression by means of
S-estimators. In Robust and Nonlinear Time Series Analysis. Lecture Notes
in Statist. Springer, New York. 26 256-272

\bibitem{S81} Stahel, W.\ A.\ (1981), Robuste Schatzungen: Infinitesimale Optimalitiit und Schiitzungen von
 Kovarianzmatrizen. Ph.D. dissertation, ETH, Zurich.
\bibitem{SZF20}
Sun, Q., Zhou, W.X., and Fan, J.Q. (2020), ``Adaptive Huber
Regression", {\it Journal of the American Statistical Association}, 115:529, 254-265, DOI:
10.1080/01621459.2018.1543124

\bibitem{W09} Weber, A., 1909, ``Uber den Standort der Industrien, Tubingen", In: Alfred Weber’s Theory of Location of Industries, University of Chicago Press. English translation by Freidrich,C.J.(1929)

\bibitem{WMZ18}Weng, H., Maleki, A., and Zheng, L, (2018), ``Overcoming the limitations of phase transition by higher order analysis of regularization techniques", {\it Ann. Statist.}  46(6A): 3099-3129.
\bibitem{WZ08} Wu, M., and Zuo, Y. (2009), ``Trimmed and Winsorized means based on a scaled deviation",
\emph{J. Statist. Plann. Inference}, 139(2), 350-365.

\bibitem{Z03} Zuo, Y. (2003) ``Projection-based depth functions and associated medians'',
\emph{Ann. Statist.}, 31, 1460-1490.

\bibitem{Z06} Y. Zuo (2006a), ``Robust location and scatter estimators in multivariate analysis'',
``The Frontiers in Statistics", Imperial College Press, 467-490.


\bibitem{Z06} Zuo, Y. (2006b), ``Multi-dimensional trimming based on projection depth", \emph{Ann. Statist.}, 34(5), 2211-2251.
\bibitem{Z18} Zuo, Y. (2018),  “A new approach for the computation of halfspace depth in high dimensions”. \emph{Communications
in Statistics - Simulation and Computation}, 48(3): 900-921.

\bibitem{ZS00a} Zuo, Y., Serfling, R., (2000), ``General notions of statistical depth function", {\it Ann. Statist.}, 28, 461-482.
\end{thebibliography}
}
\end{document}